\documentclass[ASNA,twocolumn]{USG} 
\usepackage{anyfontsize} %

\usepackage{colortbl}   
\usepackage{xcolor}     

\usepackage{steinmetz}

\begin{document}
\title{Gaze-Assisted Supervisory Control for Lunar Robotic Arm Teleoperation Under Concurrent Monitoring}

\author[1]{Jiayu He}
\author[1]{Yifan Li}
\author[2,3]{Oliver R. Runswick}
\author[1]{Ernest N. Kamavuako}
\author[4]{Yunjie Jia}
\author[3]{Peter D. Hodkinson}
\author[1,4,3]{Yang Gao}

\address[1]{\orgdiv{Department of Engineering, }\orgname{King's College London, }%
\orgaddress{\state{London, }\country{UK}}}

\address[2]{%
\orgdiv{Department of Psychology, Institute of Psychiatry, Psychology and Neuroscience, }%
\orgname{King's College London, }%
\orgaddress{\state{London, }\country{UK}}}

\address[3]{%
\orgdiv{Spacesuit Physiology Laboratory, Centre for Human and Applied Physiological Sciences, }%
\orgname{King's College London, }%
\orgaddress{\state{London, }\country{UK}}}

\address[4]{\orgdiv{Centre for AI Robotics in Space Sustainability, }\orgname{The Hong Kong University of Science and Technology, }%
\orgaddress{\state{Hong Kong, }\country{China}}}

\keywords{gaze-assisted supervisory control | lunar telerobotics | robotic arm teleoperation | concurrent monitoring | virtual reality}

\abstract[ABSTRACT]{Rover-mounted robotic arms can extend human reach during lunar sample handling, but continuous joystick teleoperation requires repeated visual--motor alignment while operators may also need to monitor system states. This paper presents a gaze-assisted supervisory control framework in which gaze nominates a rock or placement target, explicit confirmation authorizes the action, and the robot performs bounded autonomous grasp-and-place execution. The framework was evaluated against manual joystick teleoperation in a mission-analog virtual reality study with 16 participants. The study used a \(2 \times 2 \times 3\) within-subject design varying control mode, placement difficulty, and concurrent monitoring load. Mean recorded task completion time decreased from \(47.65\) to \(18.43\,\mathrm{s}\) per rock, with 72.5\% of this difference arising during grasping. Under the tested auditory monitoring conditions, gaze-assisted supervisory control showed less average proportional slowing, with no detected control-mode difference in monitoring-response accuracy or speed. The rate of gaze shifts between core areas of interest decreased from 49.98 to 27.73 shifts/min, while the raw NASA Task Load Index score decreased from 32.54 to 18.26. However, the benefit was task dependent. In precision placement, manual control yielded lower mean placement error and a higher placement success rate. These findings support a mixed-initiative allocation in which gaze enables rapid task-level intention specification and autonomous execution, while precision-critical stages retain manual refinement.}

\maketitle

\begingroup
  \renewcommand{\thefootnote}{}
  \renewcommand{\footnotesize}{\fontsize{8}{10}\selectfont}
  \footnotetext{Corresponding author: Yang Gao (\href{mailto:yanggao@ust.hk}{yanggao@ust.hk}).}
\endgroup

\section{Introduction}\label{sec1}

Rover-mounted robotic arms can extend human operational reach, enabling operators to collect samples and deploy equipment beyond areas that can be safely or practically accessed directly~\cite{gao2017review, gao2021space}. Such support is needed because rugged terrain, abrasive dust, extreme temperatures, and severe lighting contrasts make direct crew operations hazardous and demanding~\cite{litaker2025south, NASA2025Regolith,NASA2026MoonFacts}. Nevertheless, keeping humans in the control loop remains important for interpreting task context, adjusting priorities when new samples are encountered, and responding to environmental or system uncertainties that may invalidate predefined plans~\cite{pagliara2023human,lii2024everything}.

\begin{figure}[htb]
    \centering
    \includegraphics[width=\linewidth]{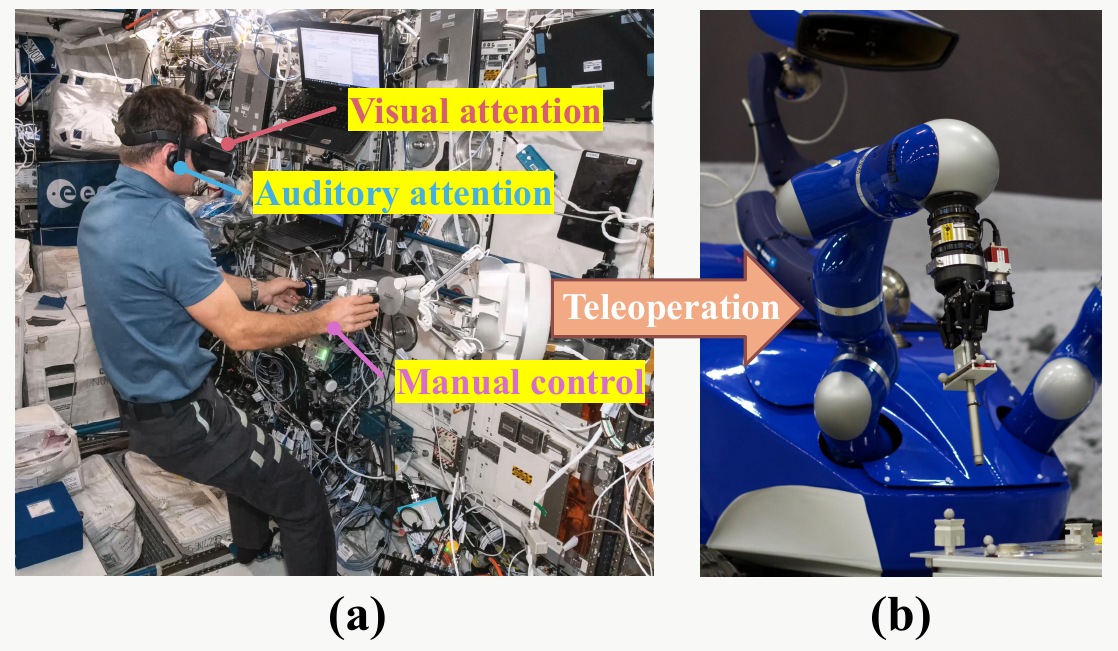}
    \caption{Conceptual illustration of manual robotic teleoperation and concurrent operator demands in a spaceflight setting. (a) An astronaut operating a robotic system using a virtual reality (VR) headset and haptic joystick, while attending to additional information, with visual, auditory, and manual demands highlighted in red, blue, and purple, respectively. (b) The teleoperated robotic arm. Source images: European Space Agency (ESA)/National Aeronautics and Space Administration (NASA), adapted from \textit{Pilotes} and \textit{Pilote timelapse} under the ESA Standard Licence.}
    \label{fig1}
\end{figure}

Manual robotic arm teleoperation typically uses a handheld controller or joystick to command end-effector motion, as illustrated in Fig.~\ref{fig1}~\cite{pagliara2023human, timman2023effect}. The operator must continuously compare the target, end-effector, and placement region and translate visual feedback into corrective manual inputs. However, limited crew availability and operational time during lunar surface missions may require operators to manage other tasks concurrently. These tasks include observing the environment, monitoring life-support and other system states through auditory or visual feedback, communicating with the crew or mission control, and confirming procedural steps, placing substantial demands on cognitive performance~\cite{anderson2025identifying}. These concurrent demands may interfere with continuous arm control because manual arm control and concurrent duties can draw on overlapping perceptual, central-processing, and response resources. Multiple Resource Theory (MRT) further suggests that presenting monitoring information acoustically may reduce modality overlap but does not eliminate competition at central-processing and response-selection stages~\cite{wickens2002multiple}. These challenges highlight the need for a control approach that retains the operator’s task-level judgment and intervention authority while reducing reliance on continuous low-level alignment and adjustment.

Supervisory control provides a way to redistribute these responsibilities. Rather than commanding each arm movement, the operator specifies what action is required and monitors its outcome, while the robot determines how to execute it~\cite{lii2024everything}. Under this allocation, the operator retains authority over target selection and command authorization and monitors autonomous execution, while autonomy is confined to bounded motion-execution stages. For lunar manipulation, this allocation allows the operator to focus on target assessment and exception monitoring, while the robot handles repeated end-effector approach, trajectory planning, and execution. By reducing the need for continuous low-level visual-motor alignment and manual input, this division may mitigate competition for perceptual, central-processing, and response resources during concurrent crew duties without removing the operator from the control loop. However, this allocation requires an efficient way for the operator to specify task-level goals without continuous manual control.

Gaze offers a potential interface for communicating task-level goals without continuous manual control. Objects or spatial regions receiving the operator’s visual attention can be identified as candidate targets~\cite{di2025collaborative}. Explicit confirmation can then distinguish intentional target selection from incidental visual attention before autonomous execution~\cite{jo2025manual}. Previous studies have explored gaze-based target selection and command input in assistive systems, wheelchairs, and mobile robots~\cite{di2025collaborative,zhang2023design}. In spaceflight research, eye tracking has primarily been used as a measurement tool to examine sensorimotor adaptation, workload, and telerobotic performance~\cite{guo2021eye}. However, gaze remains underexplored as an active task-level input for lunar surface robotic manipulation. In particular, limited evidence directly compares gaze-assisted supervisory control with continuous manual joystick teleoperation under varying placement tolerances and concurrent monitoring demands.

To address these gaps, we propose a novel gaze-assisted supervisory control framework for lunar robotic arm teleoperation. Lunar rock collection and placement are selected as a representative manipulation scenario. For this task, the proposed framework allows the operator to select the rock and placement target through gaze and explicit confirmation, while the robot is responsible for target interpretation and autonomous grasp-and-place execution. To evaluate this control allocation, we compare the framework with continuous manual joystick teleoperation in a controlled mission-analog VR study using a within-subject design that varies placement difficulty and concurrent monitoring load and assesses task performance and operator workload.

The remainder of this paper is organized as follows. Section 2 reviews related work on space telerobotics, supervisory control, and gaze-based interaction. Section 3 presents the proposed gaze-assisted supervisory control framework, while Section 4 describes the user study. Section 5 reports the experimental results, Section 6 discusses the findings and directions for future work, and Section 7 concludes the paper.

\section{Related Work}\label{sec2}

During lunar surface missions, uncertain operating conditions and constrained crew availability make it necessary for individual astronauts to manage multiple operational responsibilities, including robotic manipulation, system monitoring, environmental observation, and communication. Performing continuous manual teleoperation alongside these duties can increase workload and interfere with performance because the activities will compete for overlapping perceptual, central-processing, and response-selection resources~\cite{wickens2002multiple}. Early interface research sought to reduce teleoperation demands by improving the information presented to the operator. Predictive displays represented anticipated robot motion under communication delay~\cite{sheridan1993space}, while graphical overlays and augmented-reality aids clarified spatial relationships, planned trajectories, and collision risks~\cite{chintamani2012automated}. Later multimodal interfaces distributed information across visual, auditory, and haptic channels, with some studies reporting improved teleoperation performance~\cite{triantafyllidis2020study}. However, modality redistribution does not consistently reduce workload or concurrent-task interference because central-processing and response demands may remain~\cite{han2021investigation}. These approaches improve how operators receive information but generally leave them responsible for continuous low-level motion control.

Supervisory control can reduce this low-level control demand by reallocating selected motion-execution responsibilities from the operator to the robot. Instead of continuously commanding each end-effector movement, the operator specifies and authorizes a task-level goal and monitors its execution, while the robot performs bounded planning and motion execution~\cite{sheridan1992telerobotics}. Related shared-control approaches combine operator commands with autonomous assistance for trajectory generation, motion coordination, constraint satisfaction, and contact stabilization~\cite{li2023classification}. At the task-communication level, recent planetary-exploration research has investigated how contextual inference can support the selection of task-level commands~\cite{heikkila2012affordance}. Recent research has developed a context-aware handheld interface for commanding a rover during lunar extravehicular activity (EVA) exploration and sampling~\cite{ma2025human}. Another study used operator input and robot state to infer subtask goals and recommend task-level commands. The approach was evaluated using Surface Avatar data from operations between the International Space Station (ISS) and the ground~\cite{mayershofer2026toward}. These studies support task-level communication through adaptive interfaces and contextual inference while retaining human authority over the task. However, supervisory control still requires an efficient means of specifying the manipulation target and its intended destination.

Gaze offers a potential interface for specifying task-level goals because it is closely coupled with goal-directed manual action. During natural manipulation, the eyes typically fixate task-relevant objects and contact locations before the hand reaches them, supporting movement planning and control~\cite{land1999roles, johansson2001eye}. This predictive eye--hand coordination allows the attended object or spatial region to be treated as a candidate manipulation target. Beyond robotics, gaze-based interfaces have been investigated as hands-free alternatives to conventional pointing devices for selecting digital content~\cite{liao2022toward}. Gaze has also been used to select spatial features without manual cursor control in digital maps and to preselect distant menu items in augmented-reality interfaces~\cite{lystbaek2022gaze}. This input modality has also been incorporated into robot-control interfaces to specify motion directions for wheelchairs, enabling faster and more efficient navigation than selecting commands from a separate display~\cite{poy2021multimodal}. For mobile-robot teleoperation, an augmented-reality interface converted gaze position into robot motion and supported hands-free navigation~\cite{lee2023investigating}. Gaze patterns have further been used to estimate grasping and placement intentions in simulated teleoperated pick-and-place tasks~\cite{fuchs2021gaze}. Natural gaze has also been combined with joystick input for online goal prediction in shared-control teleoperation, where early predictions improved assistance~\cite{aronson2022gaze}. These studies establish the feasibility of using gaze to indicate a candidate target or direction.

However, visual attention does not necessarily indicate an intention to act because gaze is also used for scene inspection and information gathering. Jacob described this ambiguity as the Midas-touch problem: a gaze-controlled system may activate an item whenever the user looks at it, even when no command is intended~\cite{jacob1991use}. Dwell-based selection attempts to reduce accidental activation by requiring sustained fixation, but it slows selection and remains sensitive to the selected dwell threshold~\cite{paulus2021usability, mutasim2021pinch}. Multimodal interfaces instead separate gaze-based pointing from command confirmation. Button clicks and pinch gestures have been evaluated as alternatives to dwell for virtual-object selection~\cite{mutasim2021pinch}, while augmented-reality interfaces have used gaze for preselection and hand alignment to complete the selection~\cite{lystbaek2022gaze}. In wheelchair and robot teleoperation, gaze-indicated motion has also been coupled with push-button command execution~\cite{poy2021multimodal}. More recent extended reality research has evaluated dwell, swipe, tilt, and non-dominant-hand input as confirmation methods within gaze-and-pinch multiselection~\cite{kim2025pinchcatcher}. These approaches retain gaze as a rapid means of indicating a candidate target while requiring a deliberate action before execution. This separation is particularly relevant to supervisory manipulation, where gaze can indicate the object or destination and explicit confirmation can preserve operator authority over autonomous execution.

Eye-tracking research in space initially focused on physiological adaptation to microgravity. ESA-supported experiments recorded astronauts’ eye movements on the ISS and showed that weightlessness affects the coordination between balance and oculomotor control~\cite{ESA2014}. More recent research has extended eye tracking to operational human-factors assessment. In simulated Canadarm2 teleoperation, time pressure increased maximum fixation duration and pupillary activity, while latency reduced mean fixation duration and increased saccade frequency~\cite{guo2021eye}. These measures also correlated with task scores, supporting eye tracking for workload and performance assessment. In our previous work, we proposed an augmented-reality spacesuit interface that used gaze to identify objects of interest and provide contextual assistance during EVAs~\cite{he2025gaze}. Mobile eye tracking was recently tested inside analog EVA suits during locomotion, tool-cart navigation, and regolith sampling, demonstrating the feasibility of collecting and streaming stable gaze data during EVA operations~\cite{maltese2026feasibility}. These developments indicate the growing feasibility of eye tracking for astronaut assessment, operational monitoring, and interface interaction. However, limited work has evaluated gaze as a task-level input for supervisory robotic manipulation during lunar-surface EVAs. Evidence is particularly limited on whether gaze-assisted supervisory control can reduce interference from concurrent monitoring without compromising task performance, how this control allocation affects visual attention and perceived workload, and whether its effectiveness changes as placement tolerance narrows.

\section{Gaze-Assisted Supervisory Control Framework}\label{sec3}

\subsection{Framework Architecture and Human–Robot Control Allocation}\label{sec3.1}

\begin{figure*}[t]
    \centering
    \includegraphics[width=\linewidth]{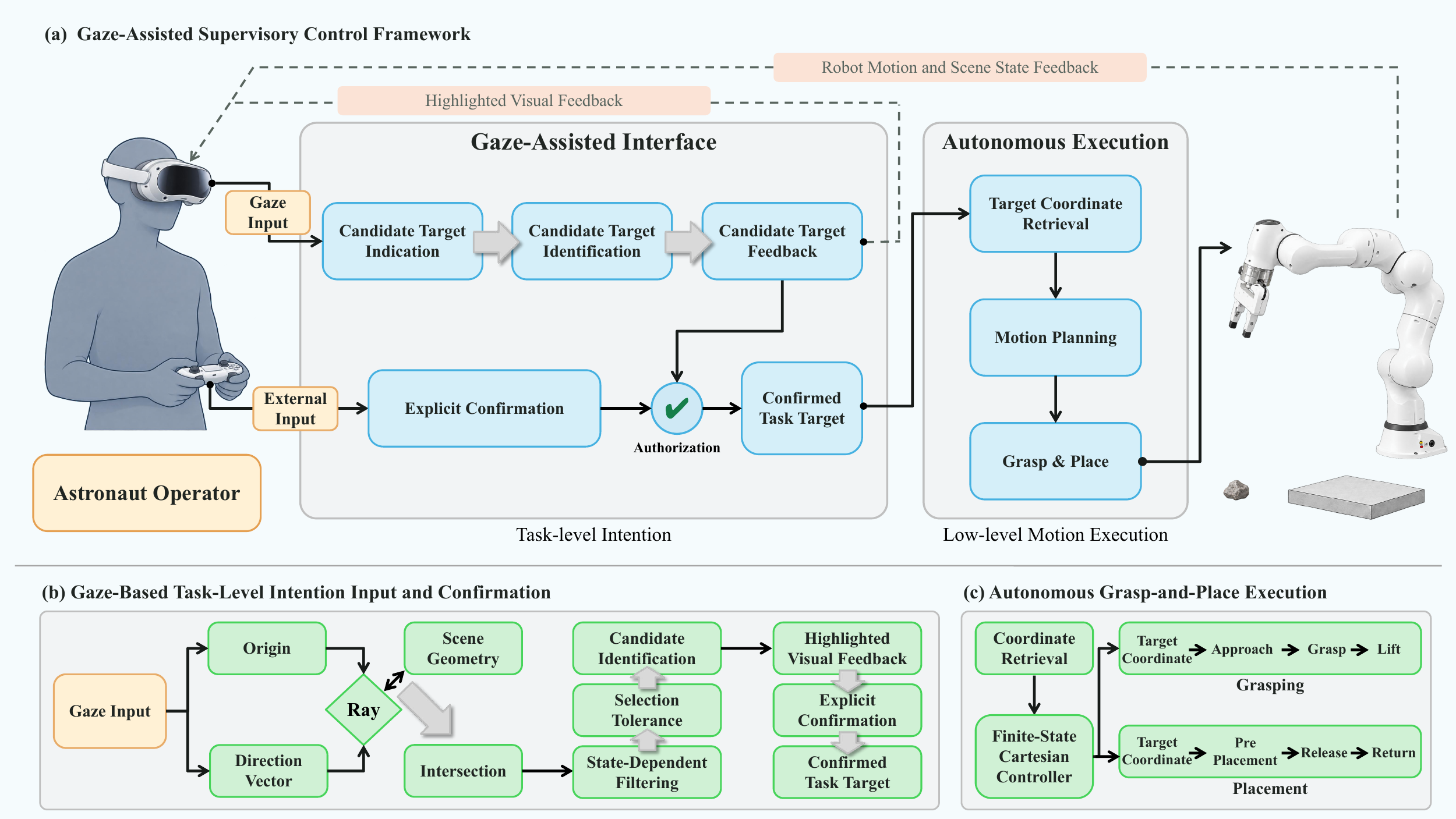}
    \caption{Overview of the proposed gaze-assisted supervisory control framework and the relationship between its interface and execution components. (a) System-level allocation of task-level intention specification and low-level robot execution. (b) Expansion of the gaze-assisted interface, showing the transformation from gaze geometry to an explicitly confirmed task target. (c) Expansion of the autonomous execution module, showing the finite-state grasping and placement sequences.}
\label{fig:framework}
\end{figure*}

To retain human task-level judgment and authorization while reducing continuous low-level manipulator control, we developed the gaze-assisted supervisory control framework illustrated in Fig.~\ref{fig:framework}~(a). The framework forms a closed human--robot control loop comprising a gaze-assisted interface, an autonomous execution module, the operator, and the robotic task environment. It separates task-level intention specification from low-level motion execution. Lunar rock collection and placement serve as the representative application of this control allocation in the present study.

The operator observes the robot and task scene and uses gaze to indicate a candidate manipulation target. Depending on the current task stage, this target represents either a rock to be grasped or a region in which the rock should be placed. The gaze-assisted interface interprets the gaze input, identifies the corresponding candidate target, and presents this candidate to the operator through visual feedback. Candidate identification alone does not initiate robot motion. Instead, the operator provides an explicit manual confirmation, which converts the candidate into a confirmed task target and authorizes the subsequent autonomous action.

Following confirmation, the autonomous execution module converts the selected target into a spatial manipulation goal. As shown in Fig.~\ref{fig:framework}~(a), this process comprises retrieving the target coordinates, planning the required manipulator motion, and executing the corresponding grasp or placement routine. Feedback on robot motion and scene state enables the operator to monitor execution and assess the task outcome. The same interaction loop is therefore applied across the manipulation sequence, while the interpretation of the confirmed target changes with the grasping or placement stage.

This architecture assigns complementary responsibilities to the operator and the robot. The operator remains responsible for interpreting the task context, deciding what should be manipulated and where it should be placed, confirming the intended action, and supervising its execution. The robot is responsible for translating the confirmed target into motion goals and performing the associated approach, alignment, grasping, transport, and placement actions. The robot does not independently determine the task goal; its autonomy is bounded by the target explicitly authorized by the operator. In contrast to continuous manual joystick teleoperation, the operator is not required to command each stage of the end-effector motion. The gaze-based target-selection and confirmation mechanism is described in Section~\ref{sec.3.2}, followed by the autonomous grasp-and-place process in Section~\ref{sec3.3}.

\subsection{Gaze-Based Task-Level Intention Input and Confirmation}\label{sec.3.2}

The gaze-assisted interface converts the operator's visual attention into a candidate task-level target. In our previous work~\cite{he2025gaze}, we investigated a gaze-conditioned object-segmentation pipeline in which the estimated gaze point was used as a point prompt for the Segment Anything Model to obtain the mask of the attended object~\cite{kirillov2023segment}. A physical implementation of the present framework could combine such gaze-conditioned segmentation with calibrated depth sensing and gaze-to-robot registration to estimate the target position in the robot workspace. In the present study, however, the framework was instantiated in a controlled VR environment, where scene geometry and target identifiers were available from the simulator. This controlled implementation allowed the effects of gaze-assisted supervisory control and human--robot control allocation to be examined independently of potential instability and errors in object segmentation.

Within this controlled VR implementation, gaze-to-target association was performed through the ray-casting pipeline illustrated in Fig.~\ref{fig:framework}~(b). At each update, the eye-tracking system provided a gaze origin, \(\mathbf{o}_{g}\), and a unit gaze-direction vector, \(\mathbf{d}_{g}\), in the headset coordinate frame. These quantities were transformed into the simulation world frame using the current headset pose. The resulting gaze ray was cast into the simulated scene and parameterized as
\begin{equation}
\mathbf{r}(\lambda)
=
\mathbf{o}_{g}
+
\lambda\mathbf{d}_{g},
\qquad \lambda \geq 0.
\end{equation}
where \(\lambda\) denotes the distance from the gaze origin along the gaze direction. The nearest intersection between this ray and the scene geometry was used to identify the scene entity under gaze. If the intersected entity was eligible under the current manipulation state, its identifier was assigned to the candidate target. The same association principle could be extended to a physical workspace by casting the gaze ray into a spatially registered and continuously updated digital twin that links modeled entities to their physical counterparts.

The task controller determines the current manipulation state according to whether the end effector is holding an object. If no object is being held, the system operates in the object-selection state, in which only uncharacterized objects of interest located on the lunar surface are eligible for selection. If the end effector is holding an object, the system operates in the placement-target-selection state, in which only eligible placement regions can be selected. Candidate-target identification is therefore restricted to the target type permitted by the current manipulation state. This state-dependent filtering initially mitigates the selection of irrelevant targets caused by transient gaze fluctuations.

To accommodate small gaze-estimation errors near target boundaries, the interface applies an ambiguity-aware selection tolerance. The tolerance margin is set to \(1.0^{\circ}\) of visual angle beyond each eligible target's projected boundary, consistent with the study's gaze-calibration acceptance criterion of less than \(1.0^{\circ}\). If the estimated gaze location directly intersects an eligible target, that target is identified as the candidate. If the gaze location instead falls on a non-selectable background region but remains within the margin of exactly one eligible target, that target can still be designated as the candidate. When two or more eligible targets fall within the same margin, the tolerance is not applied. This mechanism provides limited tolerance around isolated targets while avoiding additional ambiguity between nearby selectable targets.

The interface also uses a 100-ms look-back buffer to handle brief losses of valid gaze data. Recent valid gaze coordinates are retained within this temporal window. If the current gaze coordinate is temporarily unavailable, the interface retrieves the most recent valid coordinate from the buffer for candidate-target identification. This mechanism maintains continuity during short tracking interruptions without requiring the operator to repeat the complete target-acquisition process.

Once a valid candidate target was identified, the interface highlighted its visible surface in the headset display using a mask in the candidate-state color, allowing the operator to verify the system's interpretation. Gaze input was used only to nominate the highlighted target, and the operator confirmed the selection by pressing a controller button before autonomous execution could be initiated. This gaze-plus-button design was adopted to reduce the risk of unintended activation during ordinary visual exploration~\cite{ware1986evaluation, mutasim2022performance}.
Upon confirmation, the mask changed to the confirmed-state color. Either the object identifier for grasping or the gaze-ray intersection point for placement was then registered as the confirmed task target and transferred to the autonomous execution module. This two-stage gaze-plus-confirmation mechanism separates candidate identification from action authorization, preventing ordinary visual exploration from directly triggering robot motion while retaining operator authority over task initiation. The conversion of the confirmed target into an executable grasping or placement action is described in Section~\ref{sec3.3}.

\subsection{Autonomous Grasp-and-Place Execution}\label{sec3.3}

Autonomous grasp-and-place execution was implemented using a finite-state Cartesian controller operating at 50~Hz, with the grasping and placement sequences summarized in Fig.~\ref{fig:framework}(c). After the operator confirmed a task target, the controller generated a sequence of end-effector position targets for object approach, grasping, lifting, placement, and retreat. A motion phase was considered complete when the three-dimensional distance between the end effector and its target fell below \(0.01\,\mathrm{m}\). This finite tolerance allowed the state machine to progress without requiring exact numerical convergence to each Cartesian target.

The Cartesian targets were tracked using damped least-squares differential inverse kinematics. Given the bounded proportional position command \(\mathbf{u}_{p}\), the joint-position increment was computed as

\begin{equation}
\Delta\mathbf{q}
=
\mathbf{J}^{T}
\left(
\mathbf{J}\mathbf{J}^{T}
+
0.01^{2}\mathbf{I}_{6}
\right)^{-1}
\begin{bmatrix}
0.1\mathbf{u}_{p}\\
\mathbf{0}_{3}
\end{bmatrix},
\end{equation}

where \(\mathbf{J}\in\mathbb{R}^{6\times7}\) is the end-effector Jacobian, \(0.01\) is the damping coefficient, and \(0.1\) is the action scale. The angular component of the command remained zero, such that autonomous execution regulated the end-effector position without imposing a predefined orientation target.

For grasping, the confirmed object identifier was used to retrieve its current centre-of-mass position, \(\mathbf{p}_{o}\), from the simulator. The open gripper first approached a point \(0.05\,\mathrm{m}\) above the object,

\begin{equation}
\mathbf{p}_{\mathrm{app}}
=
\mathbf{p}_{o}
+
\begin{bmatrix}
0 & 0 & 0.05
\end{bmatrix}^{T},
\end{equation}

providing a short clearance waypoint before the final grasping motion. It then moved to \(0.01\,\mathrm{m}\) below the object centre and issued the gripper-closing command. The object was subsequently lifted \(0.30\,\mathrm{m}\) above its recorded selection position, providing clearance from the lunar surface while the robot awaited confirmation of a placement target.

For placement, the horizontal target coordinates were taken directly from the confirmed gaze-ray intersection with the placement surface. The end effector moved directly from the lifted position to \(0.05\,\mathrm{m}\) above the target surface, where the object was released. This low release clearance limited the potential free-flight distance and contact velocity of the object. Such a conservative strategy is relevant to reduced-gravity manipulation because the material and restitution properties of encountered samples may not be known a priori, making rebound, sample damage, and post-impact motion difficult to predict~\cite{brisset2018regolith,joeris2022influence}. After release, the end effector returned to the object-selection state.

\section{User Study}\label{sec4}

\subsection{Participants}\label{participants}

Sixteen participants were recruited from King’s College London, UK. The sample size was determined by an a priori power analysis using G*Power 3.1.9.7 for a within-subject repeated-measures analysis of variance with 12 measurements (\(f=0.25\), \(\alpha=0.05\), power \(=0.80\)). The sample comprised eight participants who self-reported as male and eight who self-reported as female, all aged between 21 and 29 years.
 Prior joystick experience was self-reported as none by four participants, casual by nine, frequent by two, and expert by one. Eight participants reported no use of vision correction, seven wore glasses, and one wore contact lenses. Ethical approval was granted by the Health Faculties (Purple) Research Ethics Subcommittee at King’s College London (reference number HR-25/26-54586). All participants provided written informed consent before taking part in the study.

  \begin{figure*}[t]
    \centering
    \includegraphics[width=\linewidth]{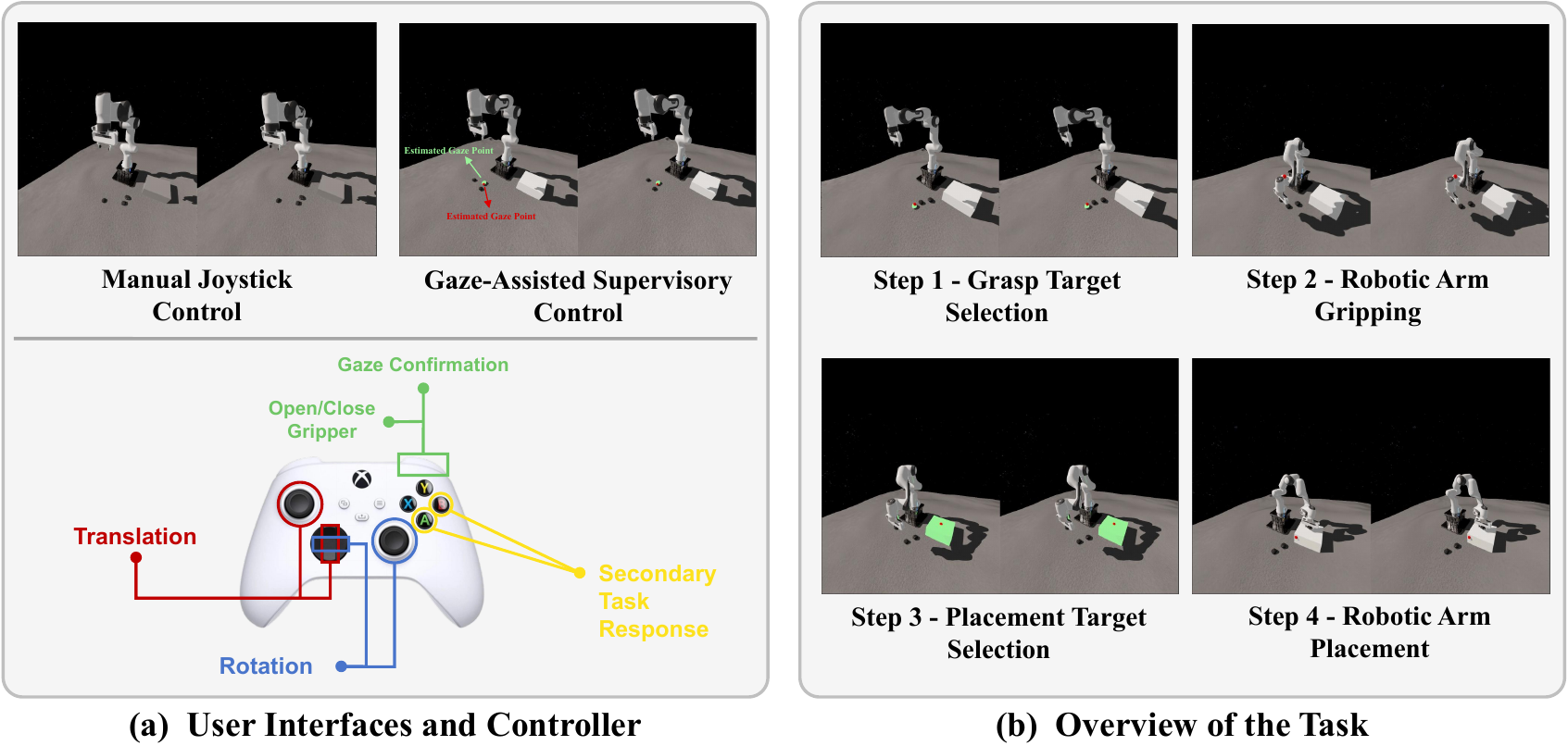}
    \caption{Binocular VR views, controller mappings, and task sequence used in the mission-analog study. (a) Paired left- and right-eye images presented through the VR headset under manual joystick teleoperation and gaze-assisted supervisory control, together with the Xbox controller mappings. In the gaze-assisted view, the red point indicates the real-time gaze position, while the identified candidate target is highlighted using a visual mask. The controller supports end-effector translation and rotation, gripper operation, explicit target confirmation, and concurrent monitoring responses. (b) Task-completion process, with gaze-assisted supervisory control used as an illustrative example. The sequence progresses through grasp-target selection, robotic-arm grasping, placement-target selection, and robotic-arm placement.}
\label{fig:vr_interface_task}
\end{figure*}

 \subsection{Experimental Setup}\label{setup}

 \subsubsection{Simulation Environment and Operator Interface}

\begin{figure}[htb]
    \centering
    \includegraphics[width=\linewidth]{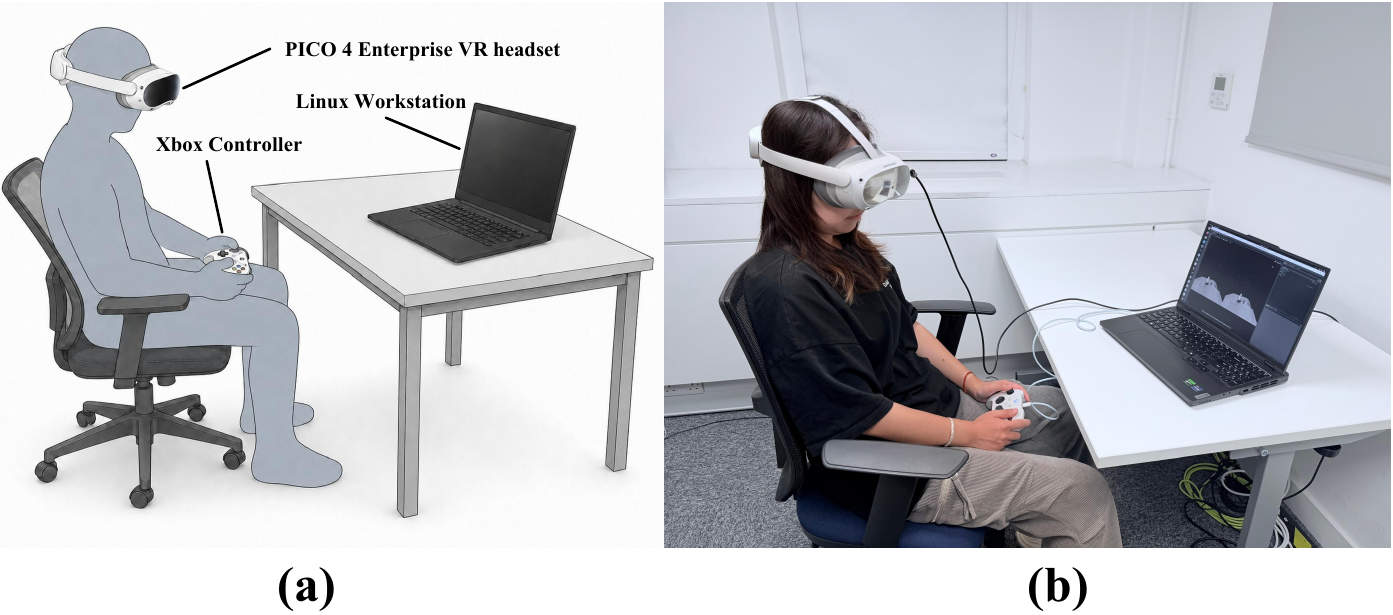}
    \caption{Experimental apparatus and operator-station configuration used in the controlled VR study. (a) Schematic of the principal hardware components, comprising a PICO 4 Enterprise VR headset with integrated eye tracking, an Xbox controller, and a Linux simulation workstation. (b) Staged demonstration of the operator-station configuration.}
    \label{fig:experimental_setup}
\end{figure}

 The study was conducted in a mission-analog VR environment developed in NVIDIA Isaac Sim by extending the Space Robotics Bench~\cite{orsula2025space} framework to support the lunar manipulation task and the operator interfaces evaluated in this study. The environment was rendered on a Linux workstation and streamed to a PICO 4 Enterprise head-mounted display. As shown in Fig.~\ref{fig:vr_interface_task}~(a), the virtual scene contained a simulated rover-mounted robotic manipulator, rock samples, and a deployment platform. The simulated manipulator was a seven-degree-of-freedom Franka arm operating under a gravitational acceleration of \(1.62\,\mathrm{m/s^2}\), corresponding to lunar surface gravity~\cite{Williams2024MoonFactSheet}. Both control modes used identical joint-velocity limits of \(2.175\,\mathrm{rad/s}\) for the first four arm joints and \(2.61\,\mathrm{rad/s}\) for the remaining three, numbered sequentially from the robot base toward the end effector. Participants viewed the environment through the headset and used an Xbox controller for manual control and explicit command confirmation. The operator-station configuration used during the study is shown in Fig.~\ref{fig:experimental_setup}.

 Gaze data were acquired using the integrated binocular eye tracker of the PICO 4 Enterprise, which supports native acquisition rates of 60 and \(90~\mathrm{Hz}\). Using OpenXR and WiVRn, we queried and logged a single unified gaze pose once per application loop at approximately \(20~\mathrm{Hz}\). This rate provided sufficient temporal resolution for the task-level gaze selections and attention measures based on areas of interest (AOIs) examined in this study.

 The operator interfaces were configured for the two control modes, as illustrated in Fig.~\ref{fig:vr_interface_task}~(a). In the manual interface, joystick inputs provided continuous end-effector motion commands, while controller buttons were used for discrete commands such as gripper operation. In the gaze-assisted interface, the integrated eye tracker was used to indicate a candidate rock or placement region, and a controller button provided explicit confirmation. The same headset and controller were used in both modes, while the primary input changed from continuous joystick commands to gaze-based target indication with button confirmation.

 \subsubsection{Robotic Task and Control Conditions}

 In each trial, participants used the rover-mounted robotic manipulator to collect three rock samples from the simulated lunar surface and transfer them to the deployment platform. For each rock, the task comprised target acquisition, approach, grasping, transport, and placement. This sequence was repeated until all three rocks had been placed.

\begin{figure}[htb]
    \centering
    \includegraphics[width=\linewidth]{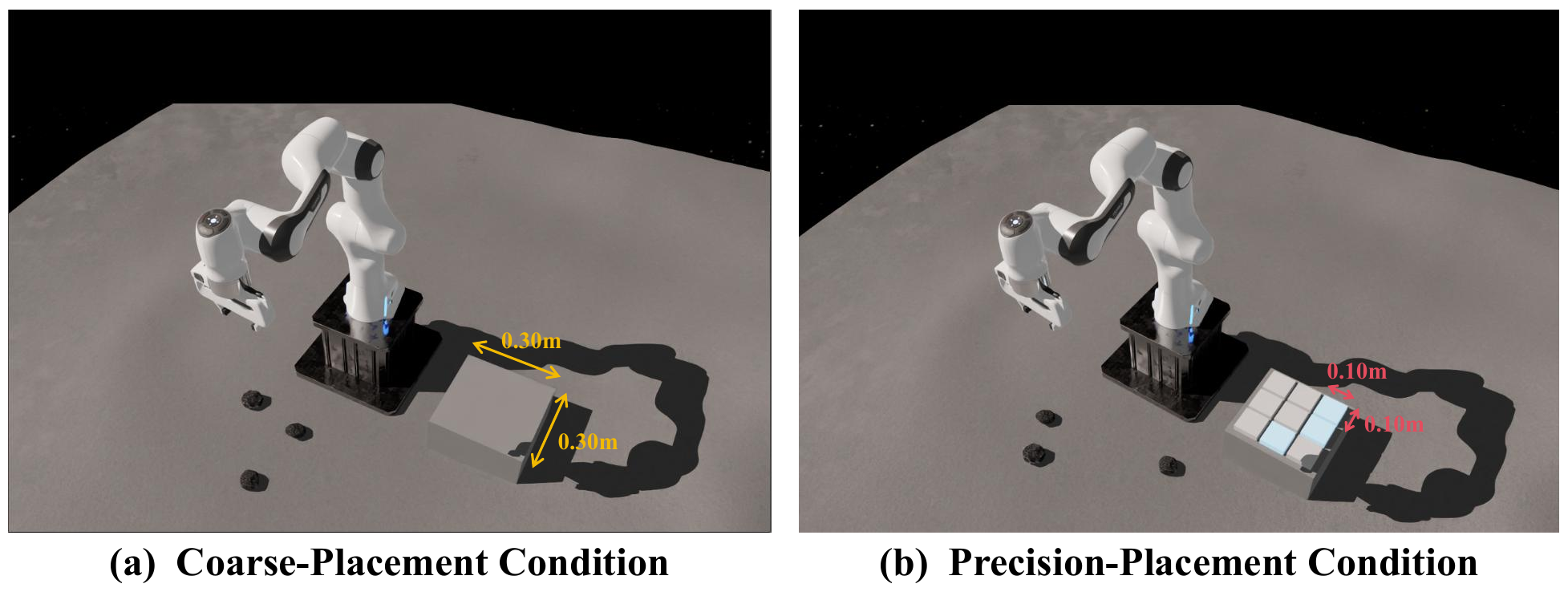}
    \caption{Placement-platform configurations used to manipulate placement difficulty in the VR study. The highlighted grid cell indicates the designated target region in the precision-placement condition.}
    \label{fig:placement_conditions}
\end{figure}

Two placement-difficulty conditions (Fig.~\ref{fig:placement_conditions}) were evaluated by varying the size of the allowable target region while maintaining the same nominal movement amplitude, following the target-width principle of Fitts’ law~\cite{fitts1954information}. The three rocks were initially positioned at randomly generated locations within a \(0.25 \times 0.50\,\mathrm{m}\) region in front of the robotic manipulator, which served as the source region for grasping. The nominal centre-to-centre distance from this region to the deployment platform was approximately \(0.625\,\mathrm{m}\). In the coarse-placement condition, the allowable placement region covered the entire \(0.30 \times 0.30\,\mathrm{m}\) deployment platform. In the precision-placement condition, this region was reduced to a designated \(0.10 \times 0.10\,\mathrm{m}\) target area assigned to each rock. In precision-placement trials, the task specified the target region, and participants directed placement within that region using the assigned control mode. The two conditions therefore represented broader and more constrained endpoint-accuracy requirements, respectively.

The robotic task was performed using two control modes, manual joystick control and gaze-assisted supervisory control. Under manual joystick control, participants continuously commanded the end effector throughout the approach, grasping, transport, and placement stages and could make online corrections until the rock was released. The controller mappings used for end-effector motion and gripper operation are shown in Fig.~\ref{fig:vr_interface_task}~(a). Under gaze-assisted supervisory control, participants selected the rock and placement target through gaze and explicit button confirmation, after which the robot autonomously performed the corresponding grasping or placement sequence as described in Sections~\ref{sec.3.2} and~\ref{sec3.3}. Participants therefore retained responsibility for target selection and action authorization, while the robot assumed responsibility for low-level motion execution. The task objective and virtual environment remained the same across modes, while the interaction method and allocation of low-level motion control differed. Accordingly, the comparison evaluates the gaze-assisted supervisory-control architecture as a whole, encompassing both gaze-based target selection and autonomous motion execution.
 
 \subsubsection{Concurrent Monitoring Task}

 Concurrent monitoring load was manipulated through an auditory stimulus--response task. This allowed visual attention within the robotic task to be examined without requiring participants to redirect their gaze to a secondary display. This modality choice also distributed perceptual demands across visual and auditory resources rather than concentrating both tasks in the visual channel, consistent with MRT~\cite{wickens2002multiple}. The use of auditory communication and monitoring cues was also consistent with NASA's use of auditory communication and system-monitoring events in the Multi-Attribute Task Battery II~\cite{SantiagoEspada2011MATB}, while the oxygen and temperature alerts represented crew-facing life-support monitoring demands~\cite{NASA2026STD3001}.

Three monitoring-load conditions were evaluated. In the no-monitoring condition (L0), participants performed the robotic task without a concurrent alert-response task. In the routine-monitoring condition (L1), a `connection test'' audio cue simulated a routine communication check and required simple acknowledgement by pressing either the A or B button. In the emergency-monitoring condition (L2), either a `temperature high'' or an ``oxygen low'' alert was presented at random, requiring participants to distinguish between the alerts and press the B or A button, respectively. In L1 and L2, auditory cues occurred at random intervals of between \(8\) and \(15\,\mathrm{s}\) while participants continued the robotic task.

 \subsection{Experimental Design and Procedure}

A \(2 \times 2 \times 3\) within-subject design varied control mode, placement difficulty, and concurrent monitoring load. The respective factor levels were manual joystick control and gaze-assisted supervisory control; coarse and precision placement; and L0, L1, and L2. Each participant completed all 12 combinations, with one complete three-rock trial per condition included in the analysis. Condition order was counterbalanced across participants using a Latin-square schedule. The initial rock positions were randomized for each trial, as were the assigned target locations in the precision-placement condition and the timing and type of auditory cues in the applicable monitoring conditions.

Before the formal experiment, participants received a briefing on the robotic task, the two control modes, and the concurrent monitoring task. After the headset was fitted, the integrated eye tracker was calibrated, with calibration accepted only when the angular validation error was less than \(1.0^{\circ}\). Participants then completed structured training covering the four primary-task combinations formed by the two control modes and two placement-difficulty conditions. To meet the training criterion, they had to successfully complete one three-rock trial under each combination. The two concurrent monitoring tasks were practiced separately from robotic manipulation, with participants required to respond correctly to five cues in each condition. After satisfying these criteria, participants could request one additional practice session involving either an individual task or the combined robotic and monitoring tasks, providing further familiarization where needed while avoiding excessive practice before the formal trials.

After training, the eye tracker was recalibrated immediately before the formal experiment and again after the sixth trial, with the same validation criterion of less than \(1.0^{\circ}\) applied on both occasions. Participants then completed the 12 formal trials in their assigned counterbalanced order. All task performance, robot state, gaze, and monitoring responses were automatically recorded by the system.

After all formal trials, participants completed the NASA Task Load Index (NASA-TLX) assessments, reported their task-specific and overall control-mode preferences, and provided open-ended feedback.

 \subsection{Evaluation Measures and Statistical Analysis}\label{sec:evaluation}

\subsubsection{Task-Performance Measures}

Primary-task efficiency was assessed using task completion time (TCT), decomposed into grasping time and placement time. For each rock, timing began at the first participant-commanded movement of the end effector under manual control and when the manipulator reached the predefined waiting-for-grasp position under gaze-assisted control. In gaze-assisted control, arrival at this position preceded rock-target acquisition, gaze selection, and button confirmation, so these activities were included in grasping time. Grasping time extended from the corresponding start event to a successful grasp, while placement time covered subsequent transport and placement, including placement-target selection and confirmation under gaze-assisted control. Timing ended at completion of placement. Activities preceding the corresponding start event were not included in either mode.

The final rock position was defined as the centre-of-mass position reported by the simulator. This position was recorded after a fixed interval of 30 simulation-loop frames following the detected gripper-release event. Placement error was calculated as the horizontal Euclidean distance between this position and the centre of the corresponding target region. To account for the different target sizes, normalized placement error was calculated as

\begin{equation}
E_{\mathrm{norm}}
=
\frac{\sqrt{(x-x_{\mathrm{t}})^2+(y-y_{\mathrm{t}})^2}}{W},
\end{equation}

where \((x,y)\) is the recorded final horizontal position of the rock’s centre of mass, \((x_{\mathrm{t}},y_{\mathrm{t}})\) is the target-region centre, and \(W\) is the full width of the allowable placement region: \(0.30\,\mathrm{m}\) for coarse placement and \(0.10\,\mathrm{m}\) for precision placement. Placement success required the centre-of-mass position to satisfy the horizontal acceptance bounds and the corresponding platform-height criterion. Attempts that could not be completed because the manipulator reached a joint limit were terminated and repeated. For each participant and condition, the last complete three-rock trial was retained for analysis. Unsuccessful placements within retained trials, including rocks placed outside the required target region, remained in both the placement-error and success analyses.

Concurrent monitoring effects on TCT were additionally expressed as dual-task cost (DTC):

\begin{equation}
\mathrm{DTC}_{\mathrm{TCT}}
=
\frac{\mathrm{TCT}_{L_k}-\mathrm{TCT}_{L_0}}
{\mathrm{TCT}_{L_0}}
\times 100\%,
\qquad k\in\{1,2\},
\end{equation}

where \(\mathrm{DTC}_{\mathrm{TCT}}\) denotes the DTC expressed as the percentage change in TCT, and \(\mathrm{TCT}_{L_k}\) and \(\mathrm{TCT}_{L_0}\) denote the TCT values under concurrent-monitoring condition \(L_k\) and the no-monitoring condition, respectively. Positive values indicated increased completion time under concurrent monitoring. Monitoring-task performance was assessed using response accuracy and response time from cue onset to the registered controller response.

\subsubsection{Visual-Attention and Subjective Measures}

Valid gaze samples were assigned to four AOIs comprising the target rock, end effector, deployment platform, and other scene regions using the three-dimensional gaze-intersection point described in Section~\ref{setup}. Visual-attention measures included AOI transition count and rate, mean continuous core-AOI dwell duration, and the proportion of valid gaze time directed toward each AOI. Transition rates were expressed per minute of valid gaze to account for differences in task duration between control modes. Invalid gaze samples and samples assigned to other scene regions terminated continuous core-AOI dwell segments, and no core-AOI transitions were counted across these interruptions.

After all 12 formal trials, participants completed one retrospective NASA-TLX assessment for each control mode. The raw NASA-TLX score was the unweighted arithmetic mean of the six subscale ratings on a 0--100 scale. The performance subscale was oriented so that lower values indicated better self-rated performance. Task-specific and overall control-mode preferences were also collected.

\subsubsection{Statistical Analysis}

Overall control-mode differences in TCT, core-AOI transition rate, continuous core-AOI dwell duration, and raw NASA-TLX scores were assessed using two-sided paired \(t\)-tests on participant-level summaries. Mixed-effects models of log-transformed TCT and per-rock placement error examined control-mode effects and their dependence on monitoring load or placement difficulty. Monitoring interference was assessed through TCT-model interactions and contrasts relative to L0, supplemented by descriptive DTC summaries. Precision-placement failure was analysed using participant-clustered generalized estimating equations (GEE). Monitoring correctness was analysed using binomial GEE, while correct positive response times were analysed on the logarithmic scale. Core-AOI transition counts were analysed using negative-binomial GEE with valid-gaze duration as the exposure.

Model effects were assessed using Wald tests, with contrasts reported alongside 95\% confidence intervals (CIs). Holm correction was applied within five comparison families: the four overall paired comparisons (TCT, core-AOI transition rate, continuous core-AOI dwell duration, and raw NASA-TLX); the six condition-specific TCT comparisons; the four DTC comparisons; the two monitoring-performance comparisons; and the two within-difficulty placement-error contrasts. Paired-test \(p\)-values shown in figure brackets are unadjusted; significance within these families was assessed using Holm-adjusted \(p\)-values at \(\alpha=.05\). Paired effect sizes were reported as Cohen's \(d_z=t/\sqrt{n}\) (\(n=16\)), using Manual minus Gaze except for NASA-TLX, which uses Gaze minus Manual. Within-mode Pearson correlations among NASA-TLX ratings were examined exploratorily, with unadjusted \(p\)-values. AOI allocation, NASA-TLX subscale ratings, and control-mode preferences were summarized descriptively.

 \section{Results}

 \subsection{Performance}

 \subsubsection{Task Completion Time}

\begin{figure*}[htb]
    \centering
    \includegraphics[width=\textwidth]{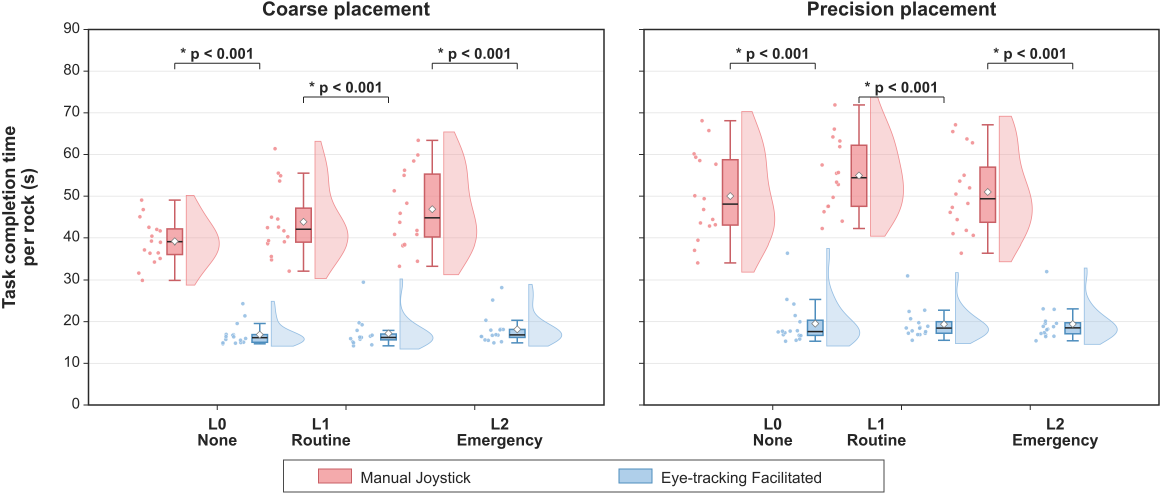}
    \caption{Per-rock TCT under manual joystick teleoperation and gaze-assisted supervisory control. Coarse and precision placement are shown separately across no (L0), routine (L1), and emergency (L2) concurrent monitoring. Each observation represents one participant's mean completion time across the three rocks in that condition; boxplots and half-violin plots show the distributions across participants. Bracketed \(p\)-values are from unadjusted two-sided paired \(t\)-tests (\(n=16\)); all six Holm-adjusted \(p\)-values were \(<.001\). Cohen's \(d_z\) values for L0, L1, and L2 were 5.258, 3.618, and 3.214 for coarse placement, and 3.681, 4.813, and 3.992 for precision placement, respectively.
}
    \label{fig:tct_by_condition}
\end{figure*}

 Figure~\ref{fig:tct_by_condition} shows lower mean recorded completion times with gaze-assisted supervisory control in all six placement-difficulty and monitoring-load conditions. Across conditions, mean recorded TCT  was 47.65~s per rock under manual joystick teleoperation and 18.43~s under gaze-assisted supervisory control. This corresponded to an average reduction of 29.22~s per rock (95\% CI [26.17, 32.27]), \(t(15)=20.433\), \(p<.001\), \(p_{\mathrm{Holm}}<.001\), \(d_z=5.108\).
 
 Most of the observed time difference occurred during grasping (Table~\ref{tab:tct_phase_decomposition}). Mean grasping time was 27.8~s per rock under manual control and 6.6~s under gaze-assisted control, while placement time was 19.9 and 11.8~s, respectively. The grasping phase accounted for approximately 72.5\% of the overall difference in TCT.
 
 The size of the time difference also varied with monitoring load. The Mode \(\times\) Load interaction was significant, Wald \(\chi^2(2)=6.293\), \(p=.043\). Relative to L0, concurrent monitoring produced a greater average proportional increase in TCT under manual control than under gaze-assisted control (ratio of load effects = 1.080, 95\% CI [1.012, 1.153], \(p=.020\)). Condition-specific DTC values are reported in Section~\ref{sec:dual_task_cost}.

 \begin{table}[htb] 
 \centering 
 \caption{Per-rock TCT by control mode and task phase. Values are means across 16 participants, averaged over placement difficulties, monitoring-load conditions, and rocks; standard deviations (SDs) are shown in parentheses.}
 \label{tab:tct_phase_decomposition} 
 \begin{tabular}{lccc} 
 \toprule Control mode 
 & \shortstack{Grasping\\time (s)} 
 & \shortstack{Placement\\time (s)} 
 & \shortstack{Total TCT(s)} \\ 
 \midrule 
 Manual joystick & 27.8 (4.6) & 19.9 (3.4) & 47.6 (7.3) \\ 
 Gaze-assisted & 6.6 (0.5) & 11.8 (3.8) & 18.4 (3.7) \\ 
 \bottomrule 
 \end{tabular} \end{table}

\begin{figure}[htb]
\centering
\includegraphics[width=\columnwidth]{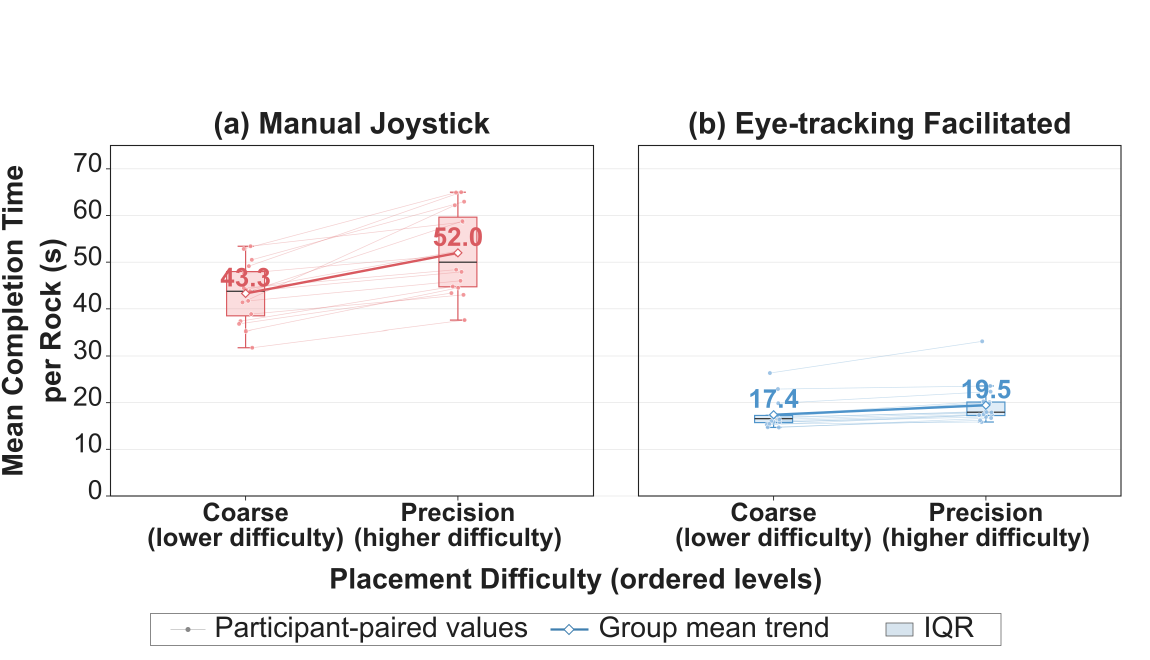}
\caption{Per-rock TCT across coarse and precision placement under (a) manual joystick teleoperation and (b) gaze-assisted supervisory control. Each point represents one participant's mean across the L0--L2 monitoring conditions, and connecting lines link the same participant's two difficulty conditions. Diamonds and thicker lines indicate group means; boxes show the interquartile ranges. The connecting lines describe paired changes between the two difficulty levels, not a continuous difficulty-response relationship.}
\label{fig:tct_by_difficulty}
\end{figure}

 When completion times were averaged across monitoring-load conditions, precision placement took longer than coarse placement for most participants under both control modes (Fig.~\ref{fig:tct_by_difficulty}). Under manual control, the observed mean increase was 8.7~s per rock, with 15 of 16 participants showing an increase. Under gaze-assisted control, the mean increase was 2.0~s per rock, with 14 of 16 participants showing an increase. Gaze-assisted control retained its time advantage at both difficulty levels.

\subsubsection{Placement Accuracy}

The effect of control mode on placement error differed between coarse and precision placement (Fig.~\ref{fig:placement_error}). A per-rock mixed model of log-transformed horizontal error showed a Mode \(\times\) Difficulty interaction, Wald \(\chi^2(1)=21.480\), \(p=3.574\times10^{-6}\). In coarse placement, mean error was 7.50~cm under manual joystick teleoperation and 6.29~cm under gaze-assisted supervisory control. The model-estimated Gaze/Manual geometric mean ratio was 0.790 (95\% CI [0.660, 0.946], Holm-adjusted \(p=.010\)). In precision placement, the direction reversed: mean error was 2.43~cm under manual control and 4.12~cm under gaze-assisted control. The model-estimated Gaze/Manual geometric mean ratio was 1.445 (95\% CI [1.206, 1.730], Holm-adjusted \(p=1.299\times10^{-4}\)). The figure displays these placement errors normalized by target-region width, whereas the model used error measured in centimetres.

All coarse placements were recorded as successful under both modes (144/144 each). In precision placement, 135/144 rocks (93.8\%) were placed successfully under manual control, compared with 126/144 (87.5\%) under gaze-assisted control. A participant-clustered GEE restricted to precision placement estimated higher odds of placement failure under gaze-assisted control (odds ratio = 2.158, 95\% CI [1.094, 4.255], \(p=.026\)).

\subsubsection{Primary-Task Dual-Task Cost} \label{sec:dual_task_cost}

\begin{figure*}[htb]
    \centering 
    \includegraphics[width=\textwidth]{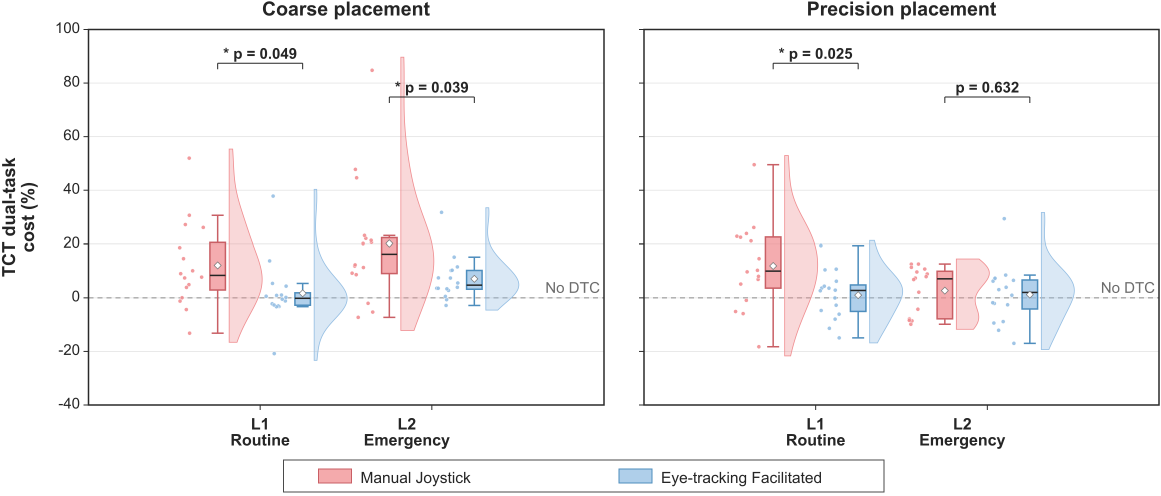}
    \caption{Primary-task DTC under manual joystick teleoperation and gaze-assisted supervisory control during coarse and precision placement. DTC represents the percentage change in TCT under concurrent monitoring relative to the corresponding no-monitoring condition (L0). Positive values indicate slower completion under concurrent monitoring, whereas negative values indicate faster completion than in L0; the dashed horizontal line denotes zero DTC. Points show individual participant values, white diamonds indicate means, and boxplots and half-violin profiles summarize the distributions. Bracketed \(p\)-values are from unadjusted two-sided paired \(t\)-tests (\(n=16\)); asterisks indicate unadjusted \(p<.05\). For coarse L1, coarse L2, precision L1, and precision L2, respectively, Holm-adjusted \(p\)-values are .118, .118, .101, and .632, and Cohen's \(d_z\) values are 0.536, 0.565, 0.622, and 0.122. These condition-specific comparisons supplement the model-based analysis of monitoring interference.} \label{fig:tct_dtc}
\end{figure*}

Primary-task DTC describes the percentage change in TCT under concurrent monitoring relative to the corresponding L0 condition for the same participant, control mode, and placement difficulty (Fig.~\ref{fig:tct_dtc}). Model-based analysis showed that the effect of concurrent monitoring differed significantly between the two control modes, as indicated by the Mode \(\times\) Load interaction, Wald \(\chi^2(2)=6.293\), \(p=.043\). Averaged across placement difficulties and the two monitoring conditions, the proportional increase from L0 was greater under manual joystick teleoperation than under gaze-assisted supervisory control (ratio of load effects \(=1.080\), 95\% CI [1.012, 1.153], \(p=.020\)). This indicates less average proportional slowing with gaze-assisted supervisory control under the auditory monitoring conditions tested. The Mode \(\times\) Difficulty \(\times\) Load interaction was not significant, Wald \(\chi^2(2)=1.733\), \(p=.420\). There was therefore no clear evidence that the Mode \(\times\) Load interaction differed between the two placement difficulties.

Descriptive DTC means were lower under gaze-assisted control in all four difficulty-by-monitoring conditions. In coarse placement, gaze-assisted control showed a consistently smaller observed DTC at both monitoring levels. Mean DTC during routine monitoring (L1) was 12.1\% under manual control and 1.7\% under gaze-assisted control. During emergency monitoring (L2), mean DTC increased to 20.2\% under manual control but remained lower at 7.1\% under gaze-assisted control. Thus, for coarse placement, completion time was less affected by concurrent monitoring with gaze assistance under both routine and emergency conditions, as reflected in the lower observed DTC values. In precision placement, the observed difference between the control modes was smaller under L2 than under L1. Under L1, mean DTC was 11.9\% for manual control and 1.0\% for gaze-assisted control, showing a markedly smaller observed cost with gaze assistance. Under L2, mean DTC was 2.7\% for manual control and 1.2\% for gaze-assisted control. Although the difference remained in the same direction, the advantage of gaze-assisted control was small in this condition. Overall, gaze-assisted supervisory control showed less interference from concurrent monitoring when the load conditions were considered together. The observed advantage was most apparent at both monitoring levels in coarse placement and under routine monitoring in precision placement. Only a limited difference between the control modes was observed during emergency monitoring in precision placement.

\begin{figure}[htb]
    \centering
    \includegraphics[width=\columnwidth]{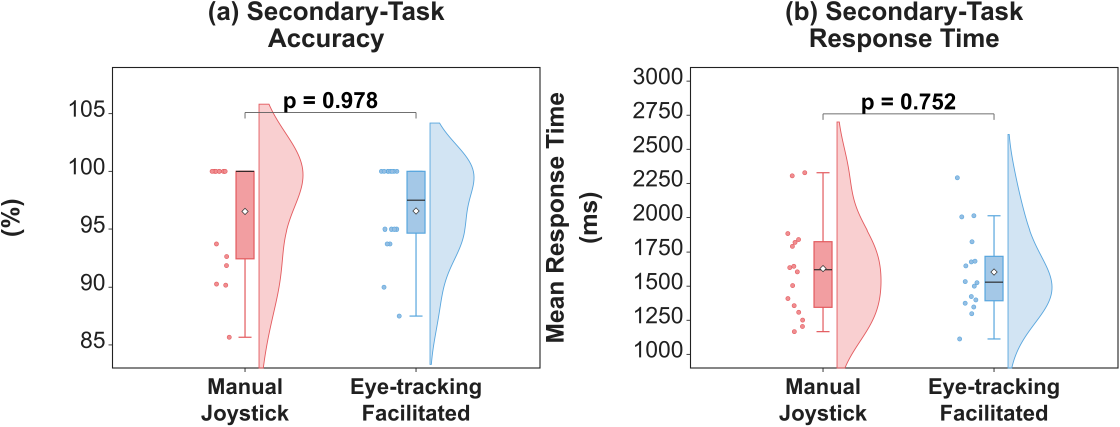}
    \caption{Secondary-task performance under manual joystick teleoperation and gaze-assisted supervisory control. (a) Monitoring-response accuracy and (b) response time for correct positive responses. L0 was excluded because no monitoring stimuli were presented in that condition. Points show participant means collapsed across placement difficulty and the L1 and L2 monitoring conditions; white diamonds indicate overall means, while boxplots and half-violin profiles summarize the distributions. Bracketed \(p\)-values are from unadjusted two-sided paired \(t\)-tests of participant-level summaries (\(n=16\)). For accuracy and response time, respectively, Cohen's \(d_z\) values are -0.007 and 0.081; both Holm-adjusted \(p\)-values are 1.000.}
    \label{fig:secondary_task_performance}
\end{figure}

No significant control-mode difference was detected in concurrent monitoring performance (Fig.~\ref{fig:secondary_task_performance}). Response correctness remained high under both control modes. The binomial GEE showed no significant effect of Mode, Wald \(\chi^2(1)=0.112\), \(p=.738\), or Load, Wald \(\chi^2(1)=0.371\), \(p=.543\), and no Mode \(\times\) Load interaction, Wald \(\chi^2(1)=0.323\), \(p=.570\).

Monitoring load affected response speed more clearly than response
correctness. Under gaze-assisted control, mean valid response time increased from 1415.9~ms under L1 to 1537.9~ms under L2. Under manual control, it increased from 1415.5~ms under L1 to 1673.4~ms under L2. The log-response-time model showed a significant increase from L1 to L2
(L2/L1 geometric mean ratio \(=1.117\), 95\% CI [1.057, 1.180];
Wald \(\chi^2(1)=15.342\), \(p<.001\)). However, the Mode \(\times\) Load interaction was not significant, Wald \(\chi^2(1)=2.673\), \(p=.102\).

Taken together, gaze-assisted supervisory control showed less average primary-task slowing under the tested auditory monitoring conditions, with no detected control-mode difference in monitoring accuracy or response speed.

\begin{figure}[htb]
    \centering
    \includegraphics[width=\columnwidth]{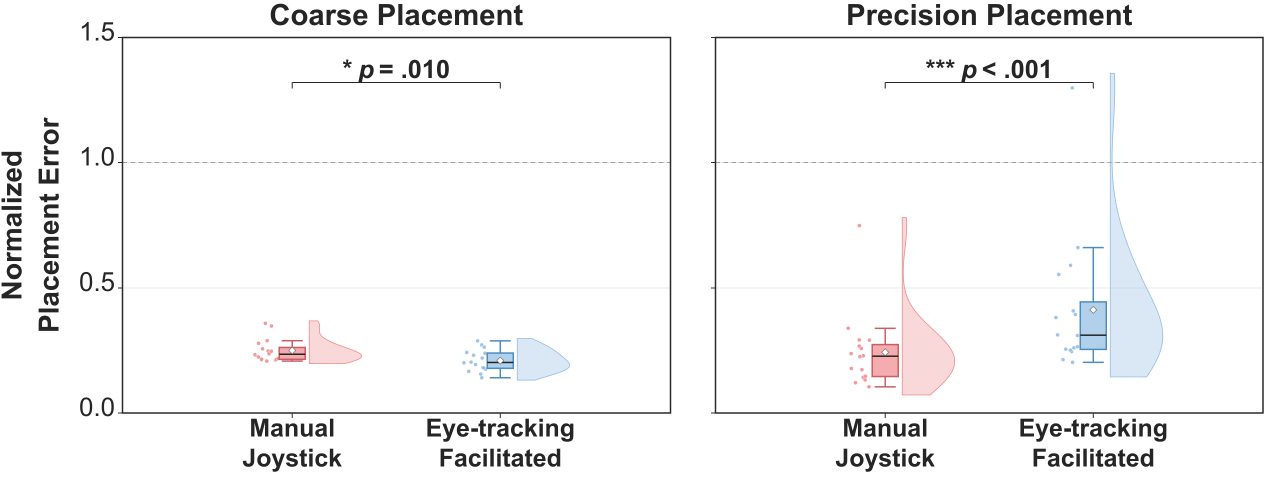}
    \caption{Normalized horizontal placement error under manual joystick teleoperation and gaze-assisted supervisory control for coarse and precision placement. Points show participant means across monitoring-load conditions and rocks; boxplots and half-violin profiles show their distributions. Lower values indicate placement closer to the target-region centre. Bracketed values are unadjusted \(p\)-values for contrasts from the placement-error mixed model; Holm-adjusted \(p\)-values are reported in the text.}
    \label{fig:placement_error}
\end{figure}

\subsection{Cognitive Load}

Gaze-assisted supervisory control was associated with lower self-reported workload than manual joystick teleoperation. The mean raw NASA-TLX score was 18.26 (SD 8.89) for gaze-assisted control and 32.54 (SD 17.70) for manual control. The paired difference was -14.27 points (95\% CI [-24.09, -4.46]), \(t(15)=-3.099\), \(p_{\mathrm{Holm}}=.015\), \(d_z=-0.775\).

Ratings for mental, physical, and temporal demand, effort, and frustration were all descriptively lower under gaze-assisted control. The performance subscale was also lower, indicating more favourable self-rated performance. These subscale patterns are descriptive; statistical inference was conducted on the overall raw NASA-TLX score.

Together with the shorter completion times and smaller average monitoring-related slowing reported above, the lower NASA-TLX score shows that the efficiency advantage of gaze-assisted control was accompanied by lower perceived workload.

\begin{figure}[htb]
    \centering 
    \includegraphics[width=\columnwidth]{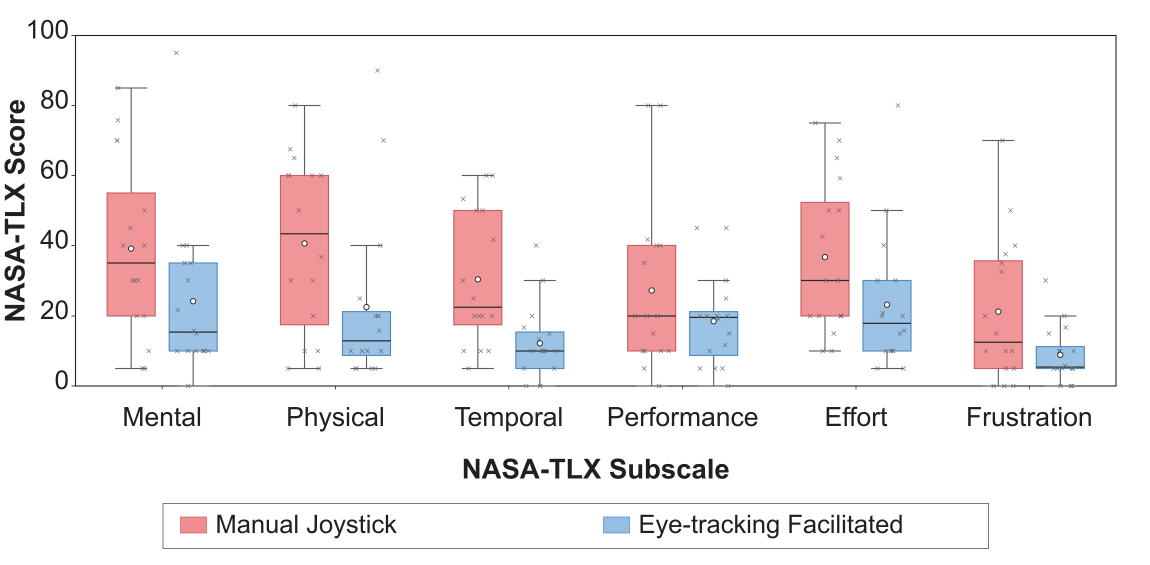} 
    \caption{Mode-level raw NASA-TLX subscale ratings for manual joystick teleoperation and gaze-assisted supervisory control (\(n=16\) per mode). Crosses show individual participant ratings, white circles indicate means, and boxplots show the distributions. Lower scores on mental, physical, and temporal demand, effort, and frustration indicate lower perceived demand. The performance subscale is scored such that lower values indicate better perceived performance. Accordingly, lower overall raw NASA-TLX scores indicate lower self-reported workload.} 
    \label{fig:nasa_tlx_subscales}
\end{figure}

\subsection{Visual-Attention}
\label{sec:visual_attention}

\subsubsection{Core-AOI Gaze Shifts and Dwell Duration}
\label{sec:gaze_shifts}

\begin{figure}[htb]
\centering
\includegraphics[width=\columnwidth]{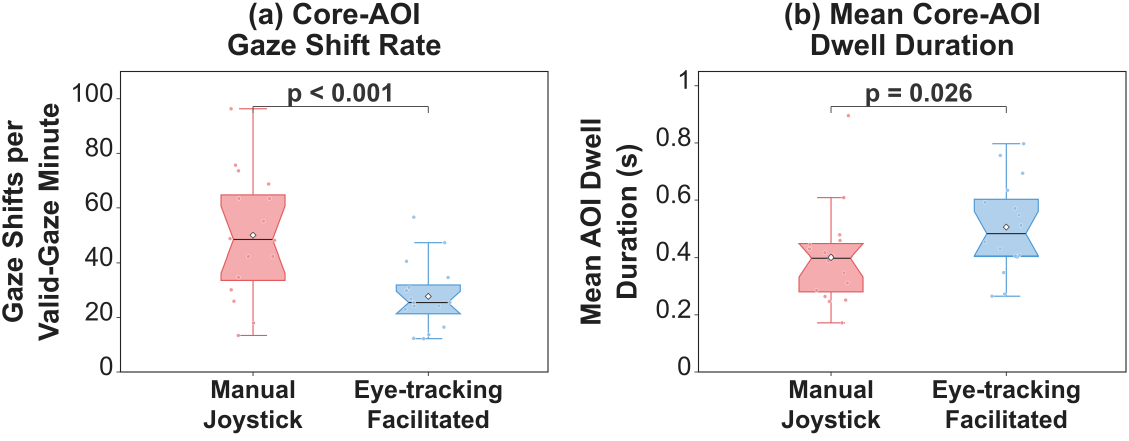}
\caption{Participant-level distributions of visual transitions and continuous dwell within core AOIs under manual joystick teleoperation and gaze-assisted supervisory control. (a) Core-AOI gaze-shift rate, calculated as transitions between consecutive valid gaze samples assigned to different core AOIs per minute of valid gaze. (b) Mean duration of consecutive samples assigned to the same core AOI. Each point represents one participant averaged across the six placement-difficulty-by-monitoring-load conditions; diamonds indicate group means, and boxplots summarize the distributions. Bracketed \(p\)-values are from unadjusted, two-sided paired \(t\)-tests (\(n=16\)). AOI dwell duration is a measure of consecutive AOI-labelled samples, not fixation duration.}
\label{fig:gaze_shifts_dwell}
\end{figure}

To account for the longer task duration under manual control, gaze shifts were expressed as transitions between core AOIs per minute of valid gaze rather than as total transition counts. As shown in Fig.~\ref{fig:gaze_shifts_dwell}(a), the participant-averaged rate was 49.98 shifts/min under manual joystick teleoperation and 27.73 shifts/min under gaze-assisted supervisory control (paired \(t\)-test, \(p<.001\), \(p_{\mathrm{Holm}}=.003\), \(d_z=1.027\)). A negative-binomial GEE analysis of condition-level transition counts, with valid-gaze duration as the exposure, likewise estimated a lower rate under gaze-assisted control (Gaze/Manual rate ratio \(=0.561\), 95\% CI [0.437, 0.720]; Wald \(\chi^2(1)=20.537\), \(p<.001\)). The model also indicated a Mode \(\times\) Difficulty interaction, Wald \(\chi^2(1)=4.200\), \(p=.040\). The relative reduction in transition rate was larger during precision placement than during coarse placement, with model-estimated Gaze/Manual rate ratios of 0.483 and 0.652, respectively.

The mean duration of consecutive gaze samples assigned to the same core AOI was 0.400~s under manual control and 0.505~s under gaze-assisted control (Fig.~\ref{fig:gaze_shifts_dwell}(b); paired \(t\)-test, \(p=.026\), \(p_{\mathrm{Holm}}=.026\), \(d_z=-0.615\)). Thus, gaze-assisted control was associated with fewer switches between core task regions and longer uninterrupted periods within a region. This dwell measure is based on consecutive AOI labels and should not be interpreted as fixation duration.

\subsubsection{Gaze Allocation Across Task Regions}
\label{sec:gaze_allocation}

\begin{figure}[htb]
\centering
\includegraphics[width=\columnwidth]{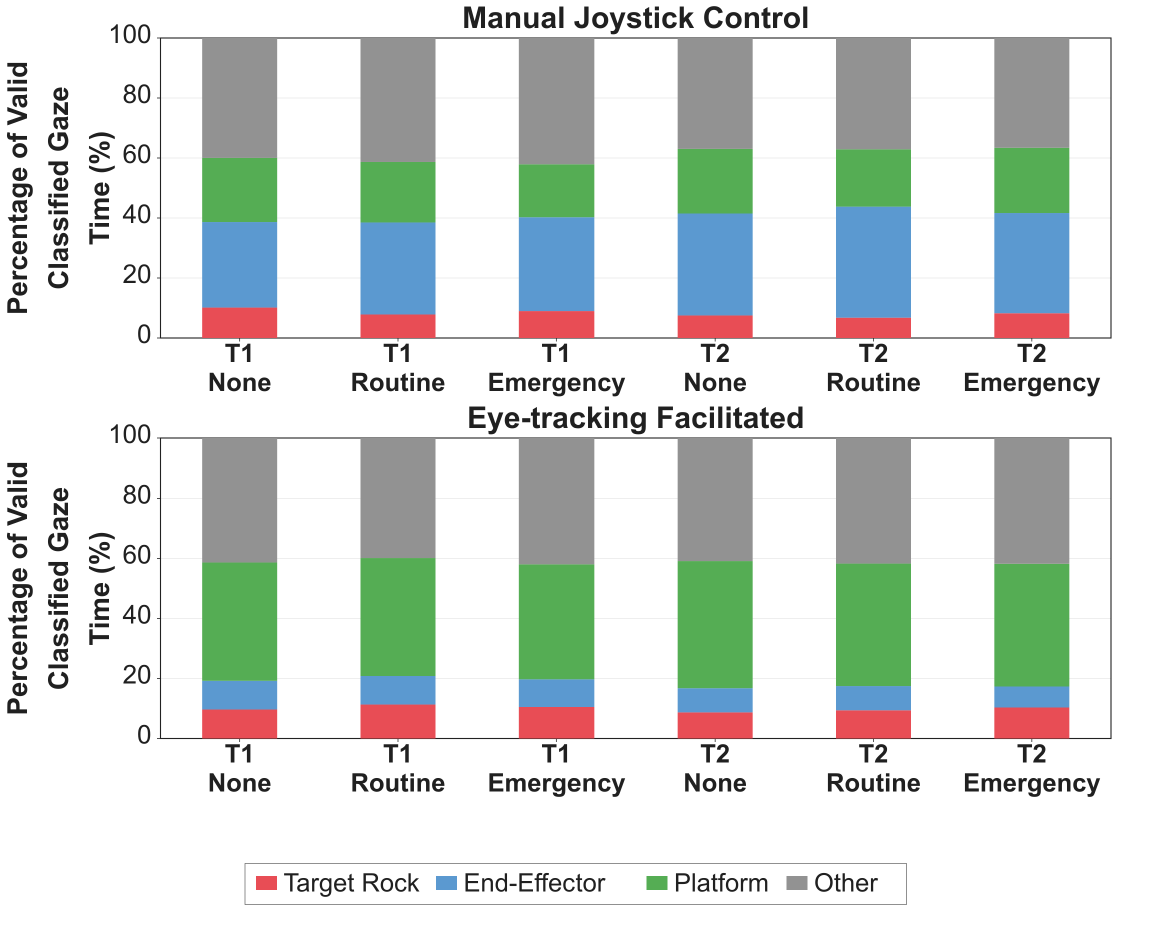}
\caption{Distribution of valid gaze time across the target rock, end effector, placement platform, and other scene regions under manual joystick teleoperation and gaze-assisted supervisory control. T1 and T2 denote coarse and precision placement, respectively; None, Routine, and Emergency denote monitoring loads L0, L1, and L2. Each stacked bar represents the mean AOI allocation across 16 participants for one placement-difficulty and monitoring-load condition. The four AOI categories are normalized to sum to 100\% within each condition. Comparisons between categories and control modes are descriptive.}
\label{fig:aoi_allocation}
\end{figure}

The two control modes also differed in where valid gaze was allocated (Fig.~\ref{fig:aoi_allocation}). Across placement difficulties and monitoring loads, manual control was associated with substantially more gaze time on the end effector than gaze-assisted control with 32.05\% versus 8.31\%. In contrast, the platform accounted for 20.05\% of valid gaze under manual control and 39.02\% under gaze-assisted control. The difference in gaze time directed toward the target rock was smaller, at 8.15\% under manual control and 9.65\% under gaze-assisted control. The remaining valid gaze was classified primarily as other scene regions.

This pattern was visible across both placement difficulties and all three monitoring-load conditions. Rather than simply increasing gaze toward the rock, gaze-assisted control was associated with less end-effector viewing and more platform viewing. Alongside the lower core-AOI transition rate, this distribution is consistent with a change from repeated monitoring of end-effector motion toward viewing the placement region during supervisory control. The AOI proportions describe logged gaze allocation; they do not, by themselves, establish a reduction in cognitive workload.

\section{Discussion and Future Work}

The mission-analog VR study shows a task-dependent benefit of the gaze-assisted supervisory-control architecture. Compared with manual joystick teleoperation, gaze-assisted supervisory control was associated with shorter recorded TCT, less average slowing under the tested auditory monitoring conditions, and lower self-reported workload scores. The placement-error advantage, however, depended on the required precision, favoring gaze-assisted supervisory control in coarse placement and manual control in precision placement. The smaller average monitoring-related increase in TCT is consistent with reduced competition between continuous arm control and concurrent response demands when low-level execution is delegated to the robot. Together with the observed monitoring performance, this pattern suggests that supervisory control can reduce manipulation-related interference under the tested concurrent-task conditions.

\begin{figure}[htp]
    \centering
    \includegraphics[width=\columnwidth]{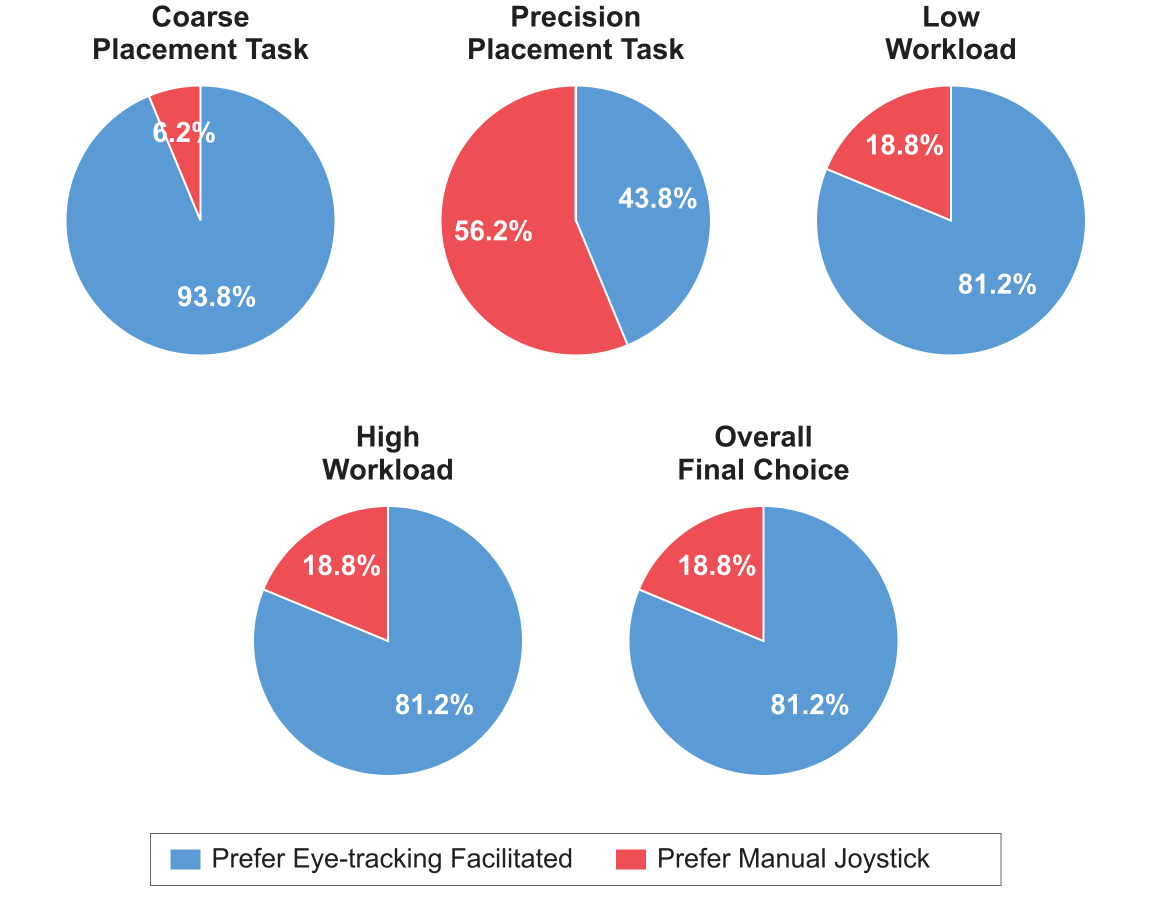}
    \caption{Participant preferences for gaze-assisted supervisory control and manual joystick teleoperation under coarse and precision placement, low and high monitoring workload, and in the overall final choice. Percentages indicate the proportion of participants preferring each mode in each context (\(N=16\)).}
    \label{fig:control_preference}
\end{figure}

The reversal in placement error suggests that the appropriate allocation of control depends on the precision required by the task. Under coarse placement, participants could select a target and delegate execution without continuously aligning the end effector. Under precision placement, the \(0.10\,\mathrm{m}\)-wide target cell was comparable in scale to the \(0.05\)--\(0.10\,\mathrm{m}\)-diameter rocks, making final positioning more constrained. Manual joystick teleoperation allowed participants to observe the approach and make small corrections before releasing the rock. In gaze-assisted supervisory control, target specification may be limited by the spatial accuracy of the gaze estimate, while placement-target generation and autonomous execution may introduce additional positioning error. Similar constraints may become more consequential in close-fitting tasks, such as inserting a connector into a socket. In such tasks, gaze could identify the intended target, while final alignment may benefit from manual adjustment or closed-loop visual and force feedback. The task-specific preferences in Fig.~\ref{fig:control_preference} are consistent with this interpretation. Participants favoured gaze-assisted control for coarse placement, while a slight majority preferred manual control for precision placement, suggesting that they recognised the value of direct adjustment as precision demands increased. This motivates a future mixed-initiative approach in which the operator retains authority to request a handover from gaze-assisted execution to manual refinement when greater precision is needed.

The observed control-mode differences also extended to how participants monitored the task visually. As reported in Section~\ref{sec:visual_attention}, gaze-assisted supervisory control involved fewer transitions between core task AOIs per minute of valid gaze and less viewing of the end effector, alongside more viewing of the placement platform. This pattern is consistent with goal-directed object manipulation, during which gaze typically precedes action and is directed toward task-relevant objects and contact locations rather than the moving hand~\cite{land1999roles,johansson2001eye}. In the present task, the greater attention to the placement region suggests a shift from tracking low-level arm motion toward supervising the outcome of autonomous execution. Delegating low-level motion execution may reduce the need to repeatedly check end-effector alignment, allowing visual monitoring to focus more on the placement goal. This redistribution of visual control demands offers a plausible explanation for the co-occurrence of fewer core-AOI transitions and lower reported workload.

\begin{figure}[htb]
    \centering
    \includegraphics[width=\columnwidth]{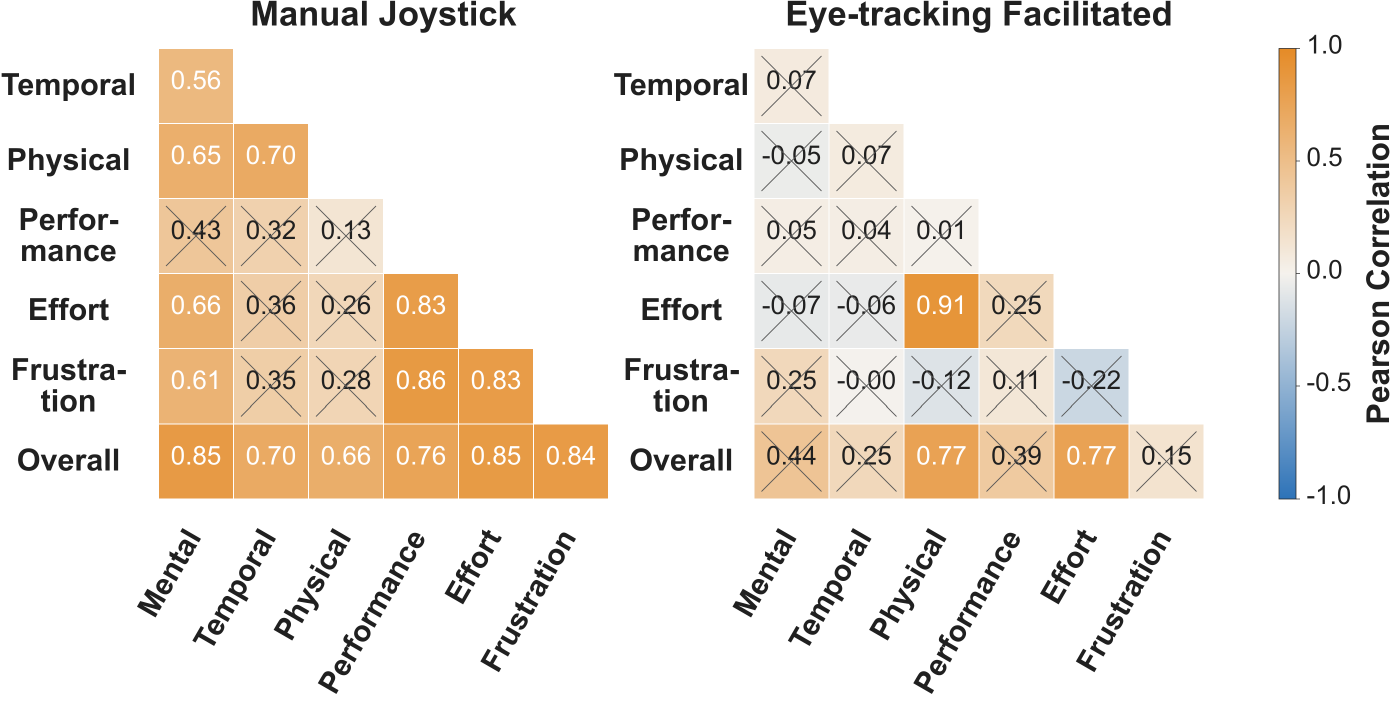}
    \caption{Exploratory within-mode Pearson correlations among NASA-TLX ratings under manual joystick teleoperation and gaze-assisted supervisory control (\(n=16\) per mode). Values and colours indicate the magnitude and direction of Pearson's \(r\), while crossed cells indicate uncorrected two-sided \(p\geq.05\).}
    \label{fig:tlx_correlations}
\end{figure}

The NASA-TLX correlations offer an exploratory, descriptive view of how different aspects of workload were related under each control mode (Fig.~\ref{fig:tlx_correlations}). Within the observed sample, manual joystick teleoperation showed a denser pattern of positive associations among workload dimensions. Gaze-assisted control showed fewer strong associations, with physical demand and effort remaining closely related. This pattern suggests broader covariation among perceived demands under manual control and a more differentiated workload profile under gaze-assisted control.

This study experimentally characterises the trade-off between task efficiency and manipulation precision under different human–robot control allocations. The results suggest that supervisory control with autonomous execution can substantially improve efficiency during repetitive manipulation, while continuous manual control remains advantageous when precise endpoint adjustments are required. These findings motivate task-dependent control allocation, where operators specify high-level goals, delegate routine manipulation to autonomous systems, and retain the ability to intervene when greater precision is needed. A specific control-allocation model and quantitative switching boundary remain topics for future investigation. The present implementation used simulator-provided geometry, object identities, and positions, and the monitoring findings are limited to the auditory conditions tested. Future work will evaluate the complete perception-to-action system in more realistic physical environments and further determine when control should transition between the two modes.

 \section{Conclusion}

This study investigated how task-level gaze input can redistribute control between the operator and robot during simulated lunar manipulation. The proposed gaze-assisted supervisory control framework used gaze and explicit confirmation to specify manipulation targets, after which the robot performed bounded autonomous grasp-and-place execution. In the mission-analog VR study, gaze-assisted control substantially shortened recorded TCT, showed less average slowing under concurrent monitoring, reduced gaze transitions among core task regions, and lowered self-reported workload compared with manual joystick teleoperation. These benefits were not accompanied by an observed reduction in concurrent monitoring accuracy or response speed. However, the effectiveness of the two control modes depended on placement precision. Gaze-assisted execution was well suited to rapid coarse manipulation, whereas manual control achieved lower placement error and a higher placement success rate in precision placement, consistent with the benefit of continuous online adjustment. These findings support task-dependent control allocation rather than selecting a single control mode for all operations. They also motivate mixed-initiative systems that combine rapid gaze-based intention specification and bounded autonomous execution with manual refinement when greater endpoint accuracy is required.

\bibliography{A_main}

\end{document}